# Immature Green Apple Detection and Sizing in Commercial Orchards using YOLOv8 and Shape Fitting Techniques

**Ranjan Sapkota, Dawood Ahmed, Martin Churuvija and Manoj Karkee**

Center for Precision and Automated Agricultural Systems, Washington State University, 24106 N Bunn Rd, Prosser, 99350, Washington, USA

Corresponding authors: ranjan.sapkota@wsu.edu and manoj.karkee@wsu.edu

**ABSTRACT** Detecting and estimating size of apples during the early stages of growth is crucial for predicting yield, pest management, and making informed decisions related to crop-load management, harvest and post-harvest logistics, and marketing. Traditional fruit size measurement methods are laborious and time-consuming. This study employs the state-of-the-art YOLOv8 object detection and instance segmentation algorithm in conjunction with geometric shape fitting techniques on 3D point cloud data to accurately determine the size of immature green apples (or fruitlet) in a commercial orchard environment. The methodology utilized two RGB-D sensors: Intel RealSense D435i and Microsoft Azure Kinect DK. Notably, the YOLOv8 instance segmentation models exhibited proficiency in immature green apple detection, with the YOLOv8m-seg model achieving the highest AP@0.5 and AP@0.75 scores of 0.94 and 0.91, respectively. Using the ellipsoid fitting technique on images from the Azure Kinect, we achieved an RMSE of 2.35 mm, MAE of 1.66 mm, MAPE of 6.15 mm, and an R-squared value of 0.9 in estimating the size of apple fruitlets. Challenges such as partial occlusion caused some error in accurately delineating and sizing green apples using the YOLOv8-based segmentation technique, particularly in fruit clusters. In a comparison with 102 outdoor samples, the size estimation technique performed better on the images acquired with Microsoft Azure Kinect than the same with Intel Realsense D435i. This superiority is evident from the metrics: the RMSE values (2.35 mm for Azure Kinect vs. 9.65 mm for Realsense D435i), MAE values (1.66 mm for Azure Kinect vs. 7.8 mm for Realsense D435i), and the R-squared values (0.9 for Azure Kinect vs. 0.77 for Realsense D435i). This study demonstrated the feasibility of accurately sizing immature green fruit in early growth stages using the combined 3D sensing and shape-fitting technique, which shows promise for improved precision agricultural operations such as optimal crop-load management in orchards.

**INDEX TERMS** YOLOv8, machine learning, deep learning, machine-vision, automation, robotics

## I. INTRODUCTION

Detecting and estimating size of apples during the early growth stages is crucial for yield prediction, pest management, harvest and post-harvest logistics, and making informed decisions related to crop-load management [1]. Accurate information on number and size of fruit during this stage enables farmers to strategize and prepare for harvest and post-harvest logistics including the workforce, equipment, and storage requirements [2]. Additionally, green fruit size serves as an indicator of tree health and vigor [3]. If the fruit does not attain its expected size during the growth stage, it could signify insufficient nutrients or pest infestation [4], [5]. By monitoring apple size and growth pattern, farmers can address these issues and improve crop production and quality to the desired level. Furthermore, accurate green fruit size estimates can help farmers predict future market reception of their crop. For instance, optimal size apples may go for higher prices than small or large apples. By knowing the size of green fruits, farmers can assess the tree's capacity to support the number of fruits and identify any potential problems that may be affecting the tree's ability to produce the desired yield of high-quality fruit [6] .

In the commercial production of tree fruit crops such as apples, farmers generally use hand tools such as measuring tape and specifically designed fruit sizer to measure the fruit size [7]. However, these techniques are labor-intensive and time-consuming, necessitating a considerable workforce in the orchard. Concurrently, commercial tree fruit growers have been grappling with labor shortages for the past two decades [8]. The COVID-19 pandemic has further exacerbated the labor shortage situation, posing a significant threat to global food security [9], [10]. Furthermore, hiring and training a seasonal or temporary workforce can be costly.

Consequently, the tree fruit industry urgently requires alternatives to human labor. The advancement in precision agriculture technologies, such as remote sensing, machine vision, and machine learning, opens the potential for the development of non-destructive, rapid methods to detect immature fruits, estimate their size and predict crop yields. Adoption of such a system can facilitate informed decisions on nutrient and water management, disease and pest control, and harvesting schedules, thereby reducing production costs and enhancing crop productivity and quality.

To realize the potential of precision agriculture in saving input while improving crop-yield and quality, the accuracy and efficacy of sensor and data processing systems is of paramount importance. Over the last decade, utilizing technological advancements, multiple state-of-the-art RGB-D sensors emerged, each with some unique capabilities. However, without a comprehensive comparison of these sensors, it remains challenging for researchers, agriculturalists, and technologists to utilize their true potential in real-world applications. Such comparisons also provide valuable information into their adaptability, and performance under varied environmental conditions.

It is also noted that despite significant advancements in machine vision and automation technologies for agricultural applications, a clear research gap exists in the automated detection and sizing of immature green apples during early growth stages. Traditional methods of apple counting and sizing primarily depend on manual measurements, which are not only time-consuming but also prone to human error. The lack of automated and accurate fruit size estimation techniques during early growth stages presents a challenge for efficient crop load management as it often leads to over- or under-thinning. Over- or under-thinning can lead to either excessive crop load, resulting in smaller fruits, or reduced yield due to excessive fruit removal. Accurate and early estimation of fruit size is also pivotal for optimizing immature green fruit thinning practices, a crucial crop-load management operation to achieve desired fruit yield and quality.

In this work, the following two objectives are pursued to support individual plant level management in tree fruit crops:

1. Develop an automated detection and sizing method for immature green fruit (fruitlet) in orchard environment using a YOLOv8-based object detection model, and sphere and ellipsoid fitting techniques.
2. Compare the performance of two widely used RGB-D sensors, the Intel RealSense 435i and the

Microsoft Azure Kinect DK, in estimating immature green apple size under complex orchard conditions.

## II. RELATED STUDY

In the past, researchers explored traditional image processing methods for fruit detection and size estimation. These conventional approaches typically involved image segmentation to separate fruits from the tree canopy, followed by the application of morphological operations. For instance, Behera et al. [11] employed color thresholding and the Randomized Hough Transform (RHT) technique to detect mangoes in canopy images. Following segmentation, an ellipsoid fitting technique was applied to delineate fruit estimate mango size in images. Similarly, Wang et al. [12] estimated mango size in trees using RGB-D images and a cascade detection method, combining 'histogram of gradients' features with Otsu's thresholding. However, this study was conducted under artificial nighttime lighting, limiting its practicality in variable natural lighting conditions.

Lin et al. [13] fused geometrical properties of a "kite" to estimate strawberry fruit size, as the kite shape resembles that of a strawberry. By segmenting calyx from strawberry fruit portions in 2D RGB images, the authors identified the boundary pixels of the fruit. However, this approach is only applicable when all fruits are equidistant from the camera. In addition, the study was conducted in a controlled environment for post-harvest grading and sorting purposes, which does not address the needs of crop load management during the growing season.

Gongal et al. [14] developed a 3D machine vision system for apple fruit size estimation in tree canopies by integrating a 2D color camera with a 3D time-of-flight (TOF) camera. They utilized histogram equalization in HSI color space to enhance color differences, followed by Otsu's thresholding and Circular Hough Transform (CHT) for apple identification. While the study reported modest accuracy (69.1% using 3D coordinates and 84.8% using 2D-pixel size), it focused on mature apples during harvest season, which cannot be used for automated crop-load management during the growing season.

Tsoulias et al. [15] recently estimated post-thinning green apple diameter using a LiDAR Laser scanner. They extracted radiometric and geometric features and applied the density-based scan algorithm (DBSCAN) to group segmented LiDAR points on the apple surface. The accuracy of this method was marginal due to particularly because of challenges posed by occlusion due to branches and leaves.

Likewise, Apolo-Apolo et al. [16] developed an unmanned aerial vehicle (UAV) and deep learning-based approach to identify, count, and estimate citrus fruit sizes on tree canopies. Using a Faster R-CNN-based method, they detected citrus fruits in aerial images captured by a DJI Phantom 3 drone. Although the study reported $R^2$ value of 0.80, the study was performed only over the parts of tree fruits that were visible in the top-view images.

Recently advanced machine learning techniques have been extensively applied for fruit size estimation in agriculture [17]. For example, Omeed et al. [18] used a Convolutional Neural Network (CNN) model for on-tree kiwifruit detection and size estimation. Li et al. [19] employed a Random Forest algorithm to estimate matured apple size using 3D images captured with a structured light-based imaging system. Fu et al. [20] utilized a Faster R-CNN-based model to detect and segment matured apples in RGB images and applied a regression model for size estimation. Tobias et al. [21] recently introduced a viewpoint planning approach for identifying fruit position and size in synthetic apple images, overcoming challenges posed by occlusions from leaves. By constructing an octree with labeled fruit regions of interest, the method evaluated viewpoint candidates using a utility function that considered expected information gain, resulting in improved fruit detection and size estimation in both simulated and real-world scenarios. However, this study was limited to the glasshouse environment.

Only a few earlier studies were found for early-stage green fruit detection. Recently Wei et. al [22] developed a green fruit detection system in an orchard environment which was based on multi-scale feature extraction of target fruit by using Feature Pyramid Networks (FPN) MobileNetV2 and generated region proposal of the target fruit by using Region Proposal Network (RPN). Likewise, Gan et. al [23] proposed a system to identify immature green citrus by combining the properties of color and thermal image through a multimodal imaging platform. Additionally, for the detection of immature citrus in a dynamic environment, Li et. al[24] and Lu et. al [25] developed a fast normalized cross-correlation (FNCC) based machine vision system and Mask-RCNN network-based machine vision methods.

## III. MOTIVATION

The current state of research on fruit size estimation primarily focused on matured fruits near harvest which exhibit distinct colors compared to the background in tree canopy images. This color difference simplified the segmentation process, making it relatively easier to differentiate the fruit from the canopy. However, studies on accurately estimating fruit size during the early growth stages when fruit are small and have similar green color to other canopy parts is limited. Robotic crop load management applications require information about the number, location and size of fruits during their immature, green stages, as this data are crucial for making decisions and performing various management operations manually or with robots. For example, during fruitlet thinning, growers remove the smallest apples and retain the larger ones for future harvest. Addressing this research gap in accurately estimating fruit size during early growth stages can significantly improve the efficiency of crop load management practices, both manually and using robotic machines.

## IV. MATERIALS AND METHODS

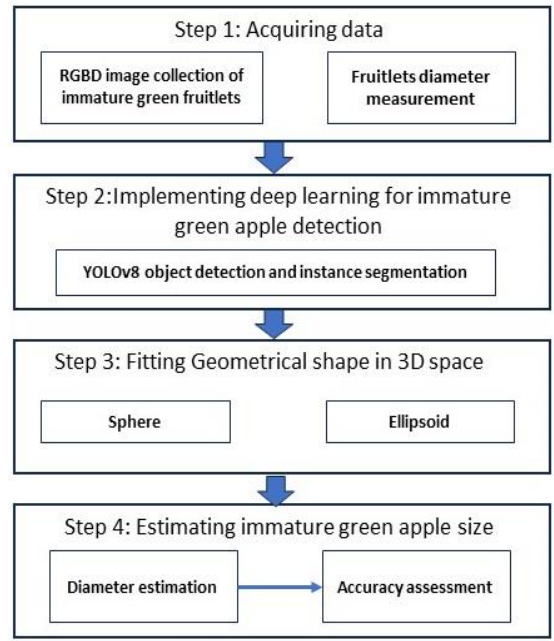

*Figure 1: Block diagram illustrating the workflow for green apple detection and size estimation.*

This study consisted of four major steps, as depicted in Figure 1, beginning with the acquisition of RGB-D data in an orchard environment, followed by early-stage green fruit detection, shape fitting, and finally fruit size estimation. The study was conducted in a commercial apple orchard with an unstructured, natural environment by employing deep learning methodologies. Fruit size estimation was achieved through the reconstruction of their corresponding geometrical shapes in 3D, utilizing sphere and ellipsoid fitting techniques. The proposed approaches were then validated in variable lighting conditions using two state-of-the-art RGB-D sensors, the Microsoft Azure Kinect (Microsoft Corporation, Redmond, USA) and the Intel RealSense 435i (Intel Corporation, California, USA).

*A. Study Site and Data Acquisition*
This study on early-stage fruit size estimation was conducted in a commercial orchard located in Prosser, Washington State, USA. The apple variety used was Scifresh and the tree rows were planted 10 feet apart with 3 feet spacing between trees. The tree height was maintained approximately at 10 feet.

In 2022, images of immature green fruits from the orchard were acquired separately from varying distances ranging from 1 to 4 feet from the tree canopy. A total of 1079 canopy images were acquired: 534 images with the Intel RealSense and 545 images with the Microsoft Azure. These images were subsequently utilized to train a deep-learning model tailored for immature green apple detection.

In 2023 season, to develop a machine vision system for potential application in real-time robotic operation in the field, the two sensors used were mounted on top of each other on a UR5e robotic arm (Universal Robots, Boston, USA) as shown in Figure 2. The robotic arm was then integrated into an unmanned ground vehicle (Warthog, Clearpath Robotics, Inc., Ontario, Canada) (Figure 2a and 2b). This setup ensured a consistent environmental condition for data acquisition with two individual sensors as the overlapping canopy images could be captured simultaneously. The cameras were aligned roughly perpendicular to the canopy and were located ~1 meter away from the target tree trunks. Precision in maintaining this specified distance was ensured using a measuring tape.

### B. Machine-Vision Sensors Used

#### 1) INTEL REALSENSE D435I

The Intel RealSense D435i is a depth-sensing camera featuring active infrared (IR) stereo vision system and an inertial measurement unit (IMU). The camera includes a 2-megapixel RGB sensor and a depth sensor with a resolution of 1280 x 720 pixels, and depth range up to 10 meters. The depth sensor operates using structured light technology, utilizing a pattern projector to create disparities between the stereo images, which are captured by two IR cameras. With a speed up to 90 frames per second (fps), and a 69.4° horizontal field-of-view (HFOV) and 42.5° vertical field-of-view (VFOV), the D435i offers flexibility for various application scenarios. The camera also includes a 6-axis IMU that provides accurate orientation data, enabling improved depth data alignment and scene understanding. The Intel RealSense D435i is compact, lightweight, and offers robust performance for depth and RGB data acquisition in a wide range of environments.

#### 2) MICROSOFT AZURE

The Azure Kinect DK sensor is equipped with a 12-megapixel color camera and a 1-megapixel depth sensor. The depth sensor operates on the Time of Flight of Light (ToF) principle and features a global shutter with analog binning, resulting in pixel-synchronized capture and reduced noise [25]. With a modulation frequency ranging from 200 to 320 MHz, the sensor offers various modes for resolution, range, and frame rate. The depth sensor has two operational depth modes: Narrow Field-of-View (NFOV) and Wide Field-of-View (WFOV). The data presented in this paper were collected using the NFOV depth mode.

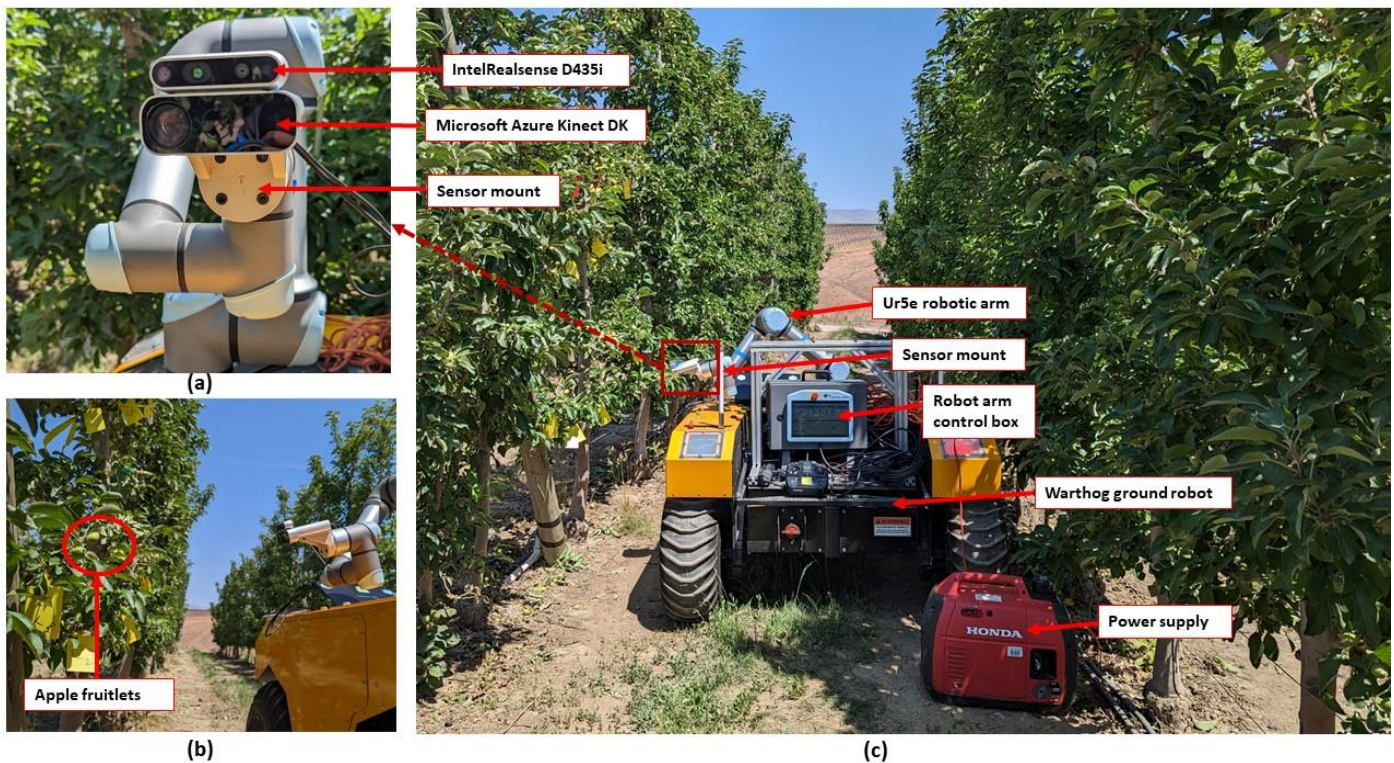

*Figure 2: Image acquisition of apple canopies with fruitlets in a commercial jazz orchard in Washington State, USA; (a) Two machine vision sensors in-action, (b) Robot and sensors facing canopy (fruitlets), and (c) Integrated system for image acquisition.*

## C. Robot Manipulation for Image Acquisition

The UR5e robotic arm was programmed such that its movement was initiated vertically upwards from the tree's base as depicted in Figure 2b while orienting the sensors to face the trees. This imaging sequence captured images from the trunk base all the way to the fifth branch layer. Throughout this maneuver, the arm consistently moved at a speed of 0.1 m/s ensuring that both sensors remained as close to perpendicular to the canopy as possible.

## D. Training YOLOv8 Model for Immature Fruit Detection

This research utilized the YOLOv8 object detection model recently provided by Ultralytics (Ultralytics, Maryland, USA; January 2023). Building upon the legacy of YOLOv5, YOLOv8 introduces enhancements that amplify its accuracy and efficiency in object detection tasks [26]. The YOLOv8 model represents a significant advancement in object detection, being anchor-free in its approach. It integrates pioneering techniques such as Pseudo Ensemble or Pseudo Supervision (PS), the adept Darknet-53 architecture, and an enhanced Feature Pyramid Network, termed YOLOv8PAFPN [27]. The principle behind PS is the concurrent training of various models, each with unique configurations, on an identical dataset. The Darknet-53 architecture, a deep 53-layer convolutional neural network, augments feature extraction, thereby enhancing object detection capabilities [28]. The anchor-free detection method emphasizes the direct estimation of an object's center instead of its relative distance from a predetermined anchor box [29] This method minimizes the number of predicted bounding boxes, simplifying the Non-Maximum Suppression (NMS) process. NMS, a demanding post-processing step, is used to filter and retain the most probable object detections after initial inference [30], [31].

The YOLOv8 model offers a suite of five distinct configurations for tasks such as identification, segmentation, and classification: YOLOv8n, YOLOv8s, YOLOv8m, YOLOv8l, and YOLOv8x. All these configurations were tested in this study for the segmentation of immature green apples from the collected images. The choice of these configurations of YOLOv8 model was based on the fundamental objective of achieving an efficient and accurate detection of immature green apples in diverse orchard environments. The images were annotated using Labelbox (Labelbox, California, USA) for training and testing the model generating 5,921 labels of immature green apples.

This labelling platform was chosen for its user-friendly interface and widespread use in the scientific community. The annotation process involved uploading images of the orchard to Labelbox, where each immature green apple was manually delineated. This delineation was achieved by drawing precise polygonal lines around the circumference of the apples, ensuring each apple was accurately segmented from the surrounding environment.

## 1. Hardware and Software

The YOLOv8 model was trained on a workstation with an Intel Xeon® W-2155 CPU @ 3.30 GHz x20 processor, NVDIA TITAN Xp Collector's edition/PCIe/SSE2 graphics card, 31.1 gigabyte memory, and Ubuntu 16.04 LTS 64-bit operating system. The training process involved setting several parameters including the learning rate, batch size, and the number of iterations to optimize the performance of the model. In this study, the learning rate was set to 0.001, while the batch size was set to 32. To prevent overfitting, a dropout rate of 0.5 was used. The model was trained for 1000 iterations while monitoring the loss function to assess the progress on model training. The images were resized to 640x640 pixels to make them compatible for YOLOv8 format.

## 2. Model Training and Testing

As mentioned above, the YOLOv8 model was trained for 1000 iterations with an early stoppage if the validation loss did not improve for 200 iterations, which provided ample opportunity for the YOLOv8 model to learn and adapt to the complexities of immature green apple segmentation in orchard environments. If the validation loss did not improve within 200 iterations, it indicated that further training would likely not yield any substantial improvements and therefore stopping the training process at that instance would improve computational efficiency and ensure the model generalizability. A batch size of 16 and an image size of 640 x 640 pixels were utilized for training. To expedite data loading and preprocessing, eight worker threads were employed in a Graphical Processing Unit (GPU). These worker threads parallelized tasks such as reading and transforming data, ensuring a consistent and efficient supply of data to the training process, and preventing unnecessary idle times in the GPU. Using multiple worker threads is beneficial in reducing data loading times, especially when dealing with large datasets.

The integration of Pseudo Supervision technique in YOLOv8 facilitates model training on a diverse range of data configurations, which allowed the model to adaptively learn intricate details of green fruits in our application. This capability not only increased the model's precision in fruit detection but also enhanced its generalization ability across different environmental conditions encountered in orchard environments.

The Stochastic Gradient Descent (SGD) optimization algorithm was employed with a learning rate of 0.01. The momentum and weight decay parameters were set to 0.937 and 0.0005, respectively, based on empirical evidence and prior research that suggested these settings offer a balance between fast convergence and model stability, reducing the chances of overfitting [32], [33]. Likewise, a warm-up phase was implemented during the first three epochs, with a warm-up momentum of 0.8 and a warm-up bias learning rate of 0.1. The incorporation of a warm-up phase during the initial training epochs served to gradually adapt the learning rate, thereby preventing the model from converging too rapidly to a suboptimal solution. More data augmentation and regularization parameters used during the training process are presented in Table 1.

*Table 1: Data augmentation and regularization parameters used in YOLOV8 training.*

| Methods Applied | Value |
|---|---|
| Hue augmentation (fraction) | 0.015 |
| Saturation augmentation (fraction) | 0.7 |
| Value augmentation (fraction) | 0.4 |
| Rotation | 0.0 |
| Translation | 0.1 |
| Scale | 0.5 |
| Flip left-right (probability) | 0.5 |
| Mosaic (probability) | 1.0 |
| Weight decay | 0.0005 |

In our specific application, we utilized this pre-trained YOLOv8 model and fine-tuned it to fit the unique requirements of the study. This fine-tuning was executed using the dataset we described above, which specifically included images of immature green apples in various orchard conditions to tailor the model's detection capabilities to our targeted agricultural setting.

*E. Performance Evaluation of YOLOv8 for Immature Fruit Segmentation*
The detection and segmentation performance of the YOLOv8 algorithm for immature green apples was evaluated using Mean Intersection over Union (MIoU), average precision (AP), mean average precision (mAP), mean average recall (mAR), and F1-score. MIoU, also known as the Jaccard index, assesses the accuracy of the segmented mask with respect to the target object, calculated as follows:

$$MIoU = \frac{Area\ Overlap}{Area\ Union} = \frac{TP}{FP + TP + FN}$$

*Equation 1*

Where,
- TP is true True Positives, which counts correctly identified apples.

- FP is False Positives, indicating non-apples mistakenly identified as apples.
- FN is False Negatives, denoting apples that were missed.

Precision evaluates the accuracy of the predicted positive detections, calculated as

$$Precision = \frac{TP}{TP + FP}$$

*Equation 2*

Recall, on the other hand, indicates how many of the actual positives our model can identify, and it is computed as:

$$Recall = \frac{TP}{TP + FN}$$

*Equation 3*

The F1-score, which considers both precision and recall, is calculated as:

$$F1 - Score = \frac{2(Precision * Recall)}{Precision + Recall}$$

*Equation 4*

AP provides a measure of the model's performance across different threshold levels, quantifying the area enclosed by the recall rate, the precision rate, and the horizontal axis. Meanwhile, mAP is a single consolidated metric to represent the model's overall detection performance. It averages the AP for all classes, providing a holistic view of the model's capability in both target detection and instance segmentation tasks.

*F. Size Estimation Using Shape Fitting in 3D Point Clouds for Segmented Immature Fruits*
Upon successful segmentation of immature green apples with the YOLOv8 model, the next vital step was the size estimation, which would be essential for various applications including growth monitoring, yield predictions, and robotic crop management operations. This process utilized point clouds, which are sets of data points in space, representing the segmented green apples. Apples in their early growth stages may have varying shapes that may start with more of a vase-shaped fruit to later resembling more like an ellipsoid and then to a sphere. To estimate accurate physical dimensions (e.g., diameter, major axis, minor axis) of fruit with these kinds of shapes, 2D color information would not be sufficient, and therefore 3D point clouds were extracted and utilized to fit two kinds of 3D shapes to represent their geometry. Open3D library (Open3D Engine, California, USA) was used to process 3D point clouds including extracting them from the raw datasets as depicted in Figure 3. This approach aimed to generate an accurate three-dimensional representation of the apples. As mentioned before, after fitting geometrical shapes to the apples delineated in the point clouds, critical physical

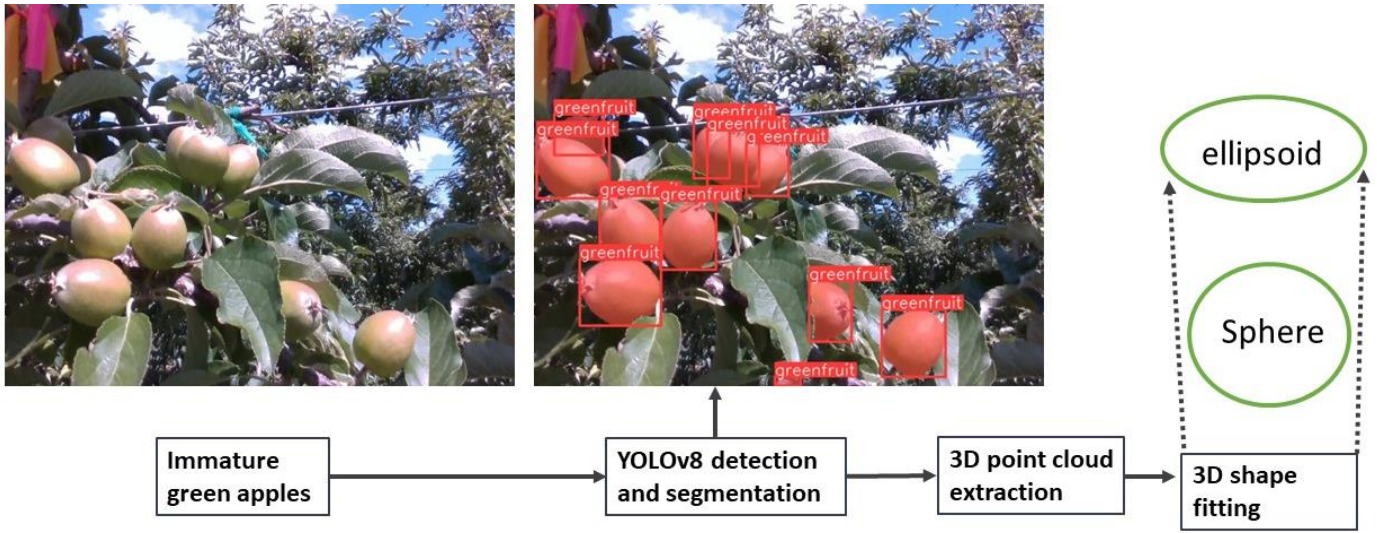

*Figure 3: Workflow diagram for YOLOv8-based green fruit segmentation and sizing. The model was trained to segment the immature green apples and their segmented mask was used to extract corresponding 3d point clouds for fitting two kinds of 3D shapes: sphere and ellipsoid.*

dimensions such as diameter, major, and minor axes of the apples were accurately determined.

Several studies in the past explored sphere fitting approach in 3D environment to estimate fruit size of peaches [34], guava[35], apples [36], pomegranate [37], tomato [38] and citrus [39]. Likewise, ellipsoid fitting technique have been explored to estimate the size of non-spherical fruits such as banana [40], watermelon [41] and mushroom [42]. Based on these past studies, the proposed study also explored the utilization of sphere and ellipsoid to reconstruct the immature green apples in 3D space using the image collected by Intel RealSense and Microsoft Azure cameras.

The ground truth measurement of immature green apples was collected using a digital caliper as shown in Figure 5b. The following shape fitting approaches were explored and compared against the ground truth:

a. Sphere Fitting Using Least Squares Method: This approach was particularly useful for immature apples that maintained a somewhat spherical shape.

b. Sphere Fitting Using Random Sample Consensus (RANSAC): While similar in aim to the previous method, this approach was robust against outliers, ensuring more accurate fittings even with occasional erroneous data points.

c. Ellipsoid Fitting: Given that not all immature green apples are perfectly spherical, the ellipsoid fitting method is catered to those apples with more elongated shapes, allowing for a better fit and thus more accurate size estimations.

### 1) SPHERE FITTING USING LEAST SQUARES
The Least Squares Fitting method focuses on the sum of squared distances from the data points to the sphere to determine the most fitting sphere that could represent the point clouds. From a set of 3D coordinates, represented as $\{p_1, p_2, ...., p_N\}$ where each point $p_i = (x_i, y_i, z_i)$, the center of the sphere, denoted as C = (a, b, c), and its radius, R, were identified to best encapsulate the immature apple. This relationship is provided by the following equation:

$$(x - a)^2 + (y - b)^2 + (z - c)^2 = R^2$$
*Equation 5*

In this equation, X = (x, y, z, R) represents the parameters of the fitted sphere: its center and radius. To ensure this sphere closely represented the immature green apple data as shown in Figure 4a, the sum of squared differences between each data point and the sphere's surface was minimized, as shown in equation 6:

$$S = \sum (f(x)^2)$$
*Equation 6*

In this context, f(x) denotes the distance between a specific data point and the modeled sphere. Through the optimization of this function, the representation of the sphere ensured to align closely with the real distribution of 3D point clouds extracted from segmented immature apples.

### 2) SPHERE FITTING USING RANSAC
The RANSAC Sphere Fitting algorithm, inherently adept at handling outliers, was implemented with a process that begins with the RANSAC algorithm choosing three random points from the dataset to hypothesize a potential sphere. These points serve to delineate the sphere's center and radius. Subsequently, all data points are cross-referenced with this hypothetical sphere. Points lying within a predefined proximity to the sphere's surface are termed 'inliers', reflecting their appropriateness to the shape, while the rest are labeled 'outliers', indicating potential errors or deviations. This process iterates, with each cycle generating a new hypothetical sphere based on a fresh trio of random points. The iteration yielding the highest inlier count identifies the most fitting sphere, whose dimensions, in turn, indicate the apple's size. As mentioned before, this method offered a rigorous approach to estimate size, especially in datasets susceptible to errors or noise. Given that the immature apples can vary in shape, exploring different shape fitting options ensured comprehensive and accurate size estimations for all samples. The mathematical representation for the RANSAC sphere fitting remains the same as the distance equation employed in the least squares method, as denoted in equations 5 and 6. Figure 4b represents the RANSAC sphere fitting over the segmented 3D point clouds of immature apples.

### 3) ELLIPSOID FITTING

For the optimal size estimation of immature green apples, especially those possessing an elongated shape, an ellipsoid fitting approach was employed to the point cloud data representation of each immature apple, aiming to estimate both the major and minor axis lengths for a detailed size profile. The ellipsoid fitting, as depicted in Figure 4c, tries to determine the most fitting ellipsoid's center, orientation, and semi-axes lengths for a collection of data points in a three-dimensional space. Mathematically, an ellipsoid is characterized by a set of points such that the sum of the squares of the distances from these points to two distinct foci remains constant. This relationship is given by equation 3.

$$\frac{(x-a)^2}{a^2} + \frac{(x-b)^2}{b^2} = 1$$

*Equation 7*

Here, (a,b) represents the ellipsoid's center, while 'a' and 'b' indicate the semi-major and semi-minor axes, respectively.

In our study, the Löwner-John ellipsoid fitting algorithm was employed. Initially, a foundational ellipsoid is derived using the least-squares method, serving as a foundation for the subsequent steps governed by the Löwner differential equation. This equation refines the ellipsoid iteratively, adapting it until the best-fitting ellipsoid for the data points is achieved, which is then used to estimate apple size. As the algorithm progresses, it fine-tunes the weights of the points and the encompassing ellipsoid's parameters. The final phase of the algorithm persists until the differences in successive weight sets become negligible, reaching a pre-set limit. Finally, this process yields a matrix 'A' and vector 'c', which encapsulate the ellipsoid's defining equation. This matrix 'A' dictates the ellipsoid's shape, while the vector 'c' pinpoints its spatial positioning.

To extract the ellipsoid's semi-axes lengths, Singular Value Decomposition (SVD) is performed on matrix 'A'. This breakdown reveals the longest (major) and shortest (minor) semi-axis lengths, crucial for the apple's size estimation. This approach provided a robust, mathematical model-based technique for estimating fruit size. This task also highlighted the importance of investigating different shape fitting techniques to cater to the diverse apple shapes encountered.

$$\frac{\partial E}{\partial t} = \left(\frac{1}{2}\right)\left(1 - \frac{(x-a)^2}{a^2} - \frac{(y-b)^2}{b^2}\right)E$$

*Equation 8*

### G. Evaluation of estimated fruit size

To evaluate the performance of the shape fitting techniques for size estimation using the two cameras, two distinct experimental setups across different years were employed. In 2022, a sample of 31 green apples with varying diameters (24, 27, 30, and 70 mm) was utilized in controlled indoor conditions, as illustrated in Figure 5a. The diverse range of synthetic green apple sizes was specifically selected to account for the variability in the field including different growth stages of green apples. Building on this, in 2023, the shape fitting techniques were further evaluated in the real-world scenarios. A comprehensive set of 102 samples was collected from a commercial jazz apple orchard environment, as had been previously detailed, with the assistance of the imaging setup controlled by a Warthog and UR5e arm.

To assess the performance in fruit sizing, Root Mean Square Error (RMSE), Mean Absolute Error (MAE), Mean Absolute Percentage Error (MAPE), and R-squared (R2) were calculated.

**Root Mean Squared Error (RMSE)**: RMSE is given by Equation 9, where n denotes the total number of observations, predicted_i corresponds to the predicted value for the ith observation, and actual_i represents the actual value for the ith observation.

$$RMSE = \sqrt{\left(\frac{1}{n}\sum(predicted_i - actual_i)^2\right)}$$

*Equation 9*

**Mean Absolute Error (MAE):** MAE calculates the average absolute difference between the estimated fruit sizes and the actual fruit sizes as given by Equation 10. It is less sensitive to outliers than RMSE, as it does not square the differences.

$$MAE = \left(\frac{1}{n}\right) * \Sigma|y_i - \hat{y}_i|$$

*Equation 10*

where:
n is the number of samples
$y_i$ is the actual size of immature fruit
$\hat{y}_i$ is the estimated size of immature fruit

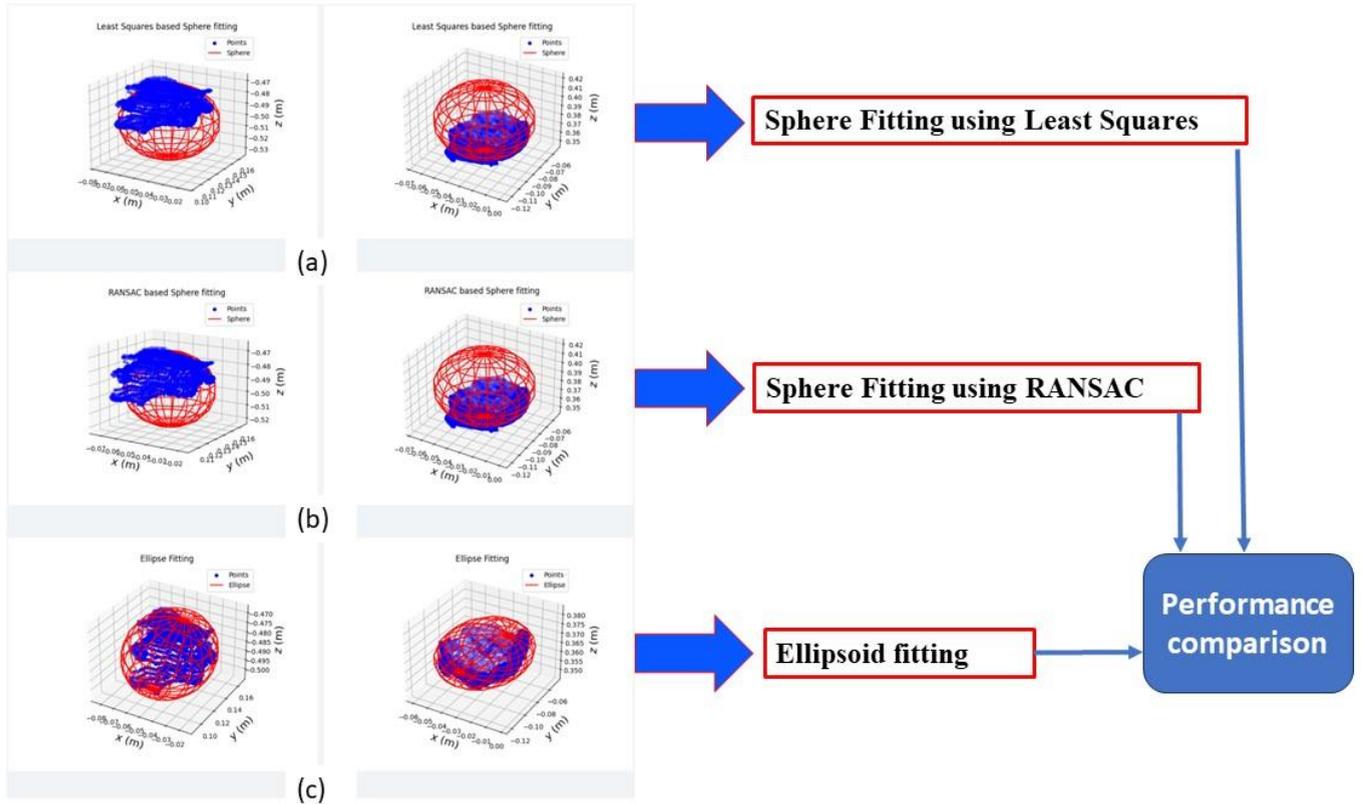

*Figure 4: 3D geometry fitting in the point clouds extracted using the segmented green apples using. YOLOv8 model; (a) Sphere Fitting using Least Squares (b); Sphere Fitting using RANSAC; (c) Ellipsoid fitting. In each image in this figure, the geometry fitting illustration shown in left side is for Realsense 435i camera and on the right is Microsoft Azure camera.*

In our case, MAE can be used to quantify the average difference in size estimation for green apples. A lower MAE indicates a better model performance.

**Mean Absolute Percentage Error (MAPE):** MAPE presents the average relative error between the estimated and the actual fruit sizes, expressed as a percentage. This allows for easier comparison across different datasets with varying scales.

$$MAPE = \left(\frac{100}{n}\right) * \Sigma \left|\frac{(y_i - \hat{y}_i)}{y_i}\right|$$

*Equation 11*

where:
 n is the number of samples
$y_i$ is the actual size of immature fruit
$\hat{y}_i$ is the estimated size of immature fruit
In your case, MAPE can be used to assess the relative performance of the size estimation workflow for green fruit. A lower MAPE indicates a better model performance.

**R-squared (R²):** R-squared is a statistical measure that represents the proportion of the variance in the dependent variable (fruit size) that can be explained by the independent variables (features used in the estimation model), which is given by.

$$R - Squared = 1 - \left(\frac{\left(\Sigma(y_i - \hat{y}_i)^2\right)}{\left(\Sigma(y_i - y_{mean})^2\right)}\right)$$

*Equation 12*

where:
$y_i$ is the actual immature fruit size
$\hat{y}_i$ is the estimated immature fruit size
$y_{mean}$ is the mean of the actual immature fruit sizes

The YOLOv8 series, particularly the YOLOv8m-seg model, has gained prominence in recent research due to its efficient single-stage architecture [1], [2], which balances model performance and computational efficiency. As indicated in our study, the YOLOv8m-seg configuration achieved the highest precision (0.9) and a commendable F1 score (0.89), outperforming other variants of YOLOv8 in terms of Average Precision at both 0.5 and 0.75 IoU thresholds. This model's balance between speed and accuracy makes it an

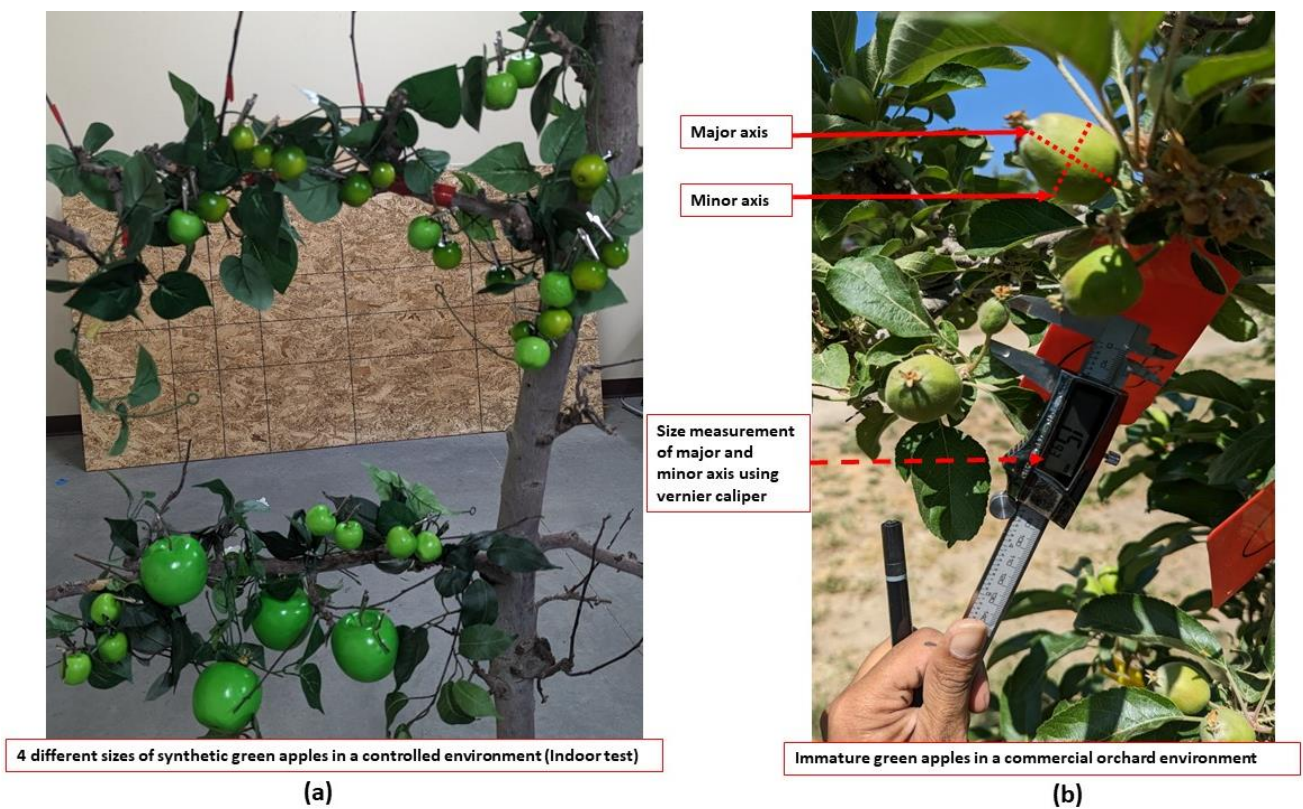

4 different sizes of synthetic green apples in a controlled environment (Indoor test)

**(a)**

Immature green apples in a commercial orchard environment

**(b)**

*Figure 5: (a) Indoor environment to estimate the size of immature green apples of different sizes, (b)Commercial orchard environment, measuring the size of immature green apples using a digital caliper.*

ideal choice for real-time applications like fruit detection in orchard environments. The selection of 3D shape fitting and sizing approaches was guided by our field observations, which show that the shape of immature apples in the early growing season would closely resemble ellipsoids, which then slowly changes to spherical shape as they grow further.

## RESULTS

### A. Immature Fruit Segmentation Using YOLOv8.

A total of 108 images containing immature apples in the natural environment of the commercial orchard were utilized to evaluate the performance of the YOLOv8 model in segmenting the immature apples. These images, captured in a commercial orchard, featured immature apples of varying sizes and orientations, providing a diverse dataset for the assessment. Figure 6 provides an example of detection results using the YOLOv8 model. Among the various configurations of the YOLOv8 algorithm tested in this study, the YOLOv8m-seg model achieved the highest precision at 0.9. The YOLOv8x-seg model marked the highest recall of 0.9. In terms of the F1 score, the YOLOv8m-seg configuration achieved the highest score of 0.89. Moreover, the YOLOv8m-seg model achieved the highest AP@0.5 of 0.94. A comprehensive breakdown of the performance metrics for all tested YOLOv8 model configurations is presented in Table 2.

YOLOv8n, being the smallest and fastest, was ideal for applications where rapid processing is essential, though with a trade-off in precision. Conversely, YOLOv8s offered a middle ground with improved accuracy while still maintaining a relatively fast processing speed. The YOLOv8m configuration offered a better balance between computational demands and segmentation accuracy. The YOLOv8m configuration made it particularly effective for complex segmentation tasks in agricultural environments. For applications where accuracy is the most critical performance measure, despite higher computational costs, larger YOLOv8 configurations such as YOLOv8l and YOLOv8x could be employed. This study focused on utilizing all these options to compare their performances for green fruit detection and to identify the most optimal model for this application.

The Precision-Recall curve, Recall-Threshold curve, F1-Score Confidence curve, and the area under the Precision-Recall curve (PR-AUC) for the YOLOv8 model are presented in Figure 7 parts a, b, c, and d, respectively. From the Precision-Recall curve, a peak precision of 0.91 was observed for green fruit detection across all classes. As can be seen in the Recall-Threshold curve, maximum recall achieved was 0.98. Similarly, PR-AUC shows a maximum value of 0.94, suggesting that a robust precision-recall performance across varied thresholds was exhibited. The model achieved the mean Average Precision (mAP) of 0.94 at an Intersection over Union (IoU) threshold of 0.5.

Although the YOLOv8 object detection and segmentation algorithm demonstrated promising capabilities with high accuracy in detecting and segmenting immature green apples in complex orchard environments with similar green color of background objects such as leaves, certain challenging situations still resulted in detection failures.

improvements could be made by training the algorithm on a larger dataset encompassing more diverse lighting conditions and varied fruit orientation to enhance its performance and robustness in addressing these challenges.

In recent years, there has been a surge in research exploring the use of YOLO-based algorithms for agricultural applications. Noteworthy among them is the study by [Sun et al.], which focused on the recognition of green apples in an orchard environment by combining the GrabCut model and Ncut algorithm [43].

*Table 2: Precision, Recall, F1-score, AP@0.5 and AP@0.75 achieved with five YOLOv8 model configurations.*

| Model | Precision | Recall | F1 Score | AP@0.5 | AP@0.75 |
|---|---|---|---|---|---|
| **YOLOv8n-seg** | 0.88 | **0.88** | 0.88 | 0.93 | 0.89 |
| **YOLOv8s-seg** | 0.89 | 0.85 | 0.87 | 0.93 | 0.89 |
| **YOLOv8m-seg** | **0.9** | **0.88** | **0.89** | **0.94** | **0.91** |
| **YOLOv8l-seg** | 0.87 | 0.87 | 0.87 | 0.93 | 0.89 |
| **YOLOv8x-seg** | 0.86 | 0.9 | 0.88 | 0.93 | 0.90 |

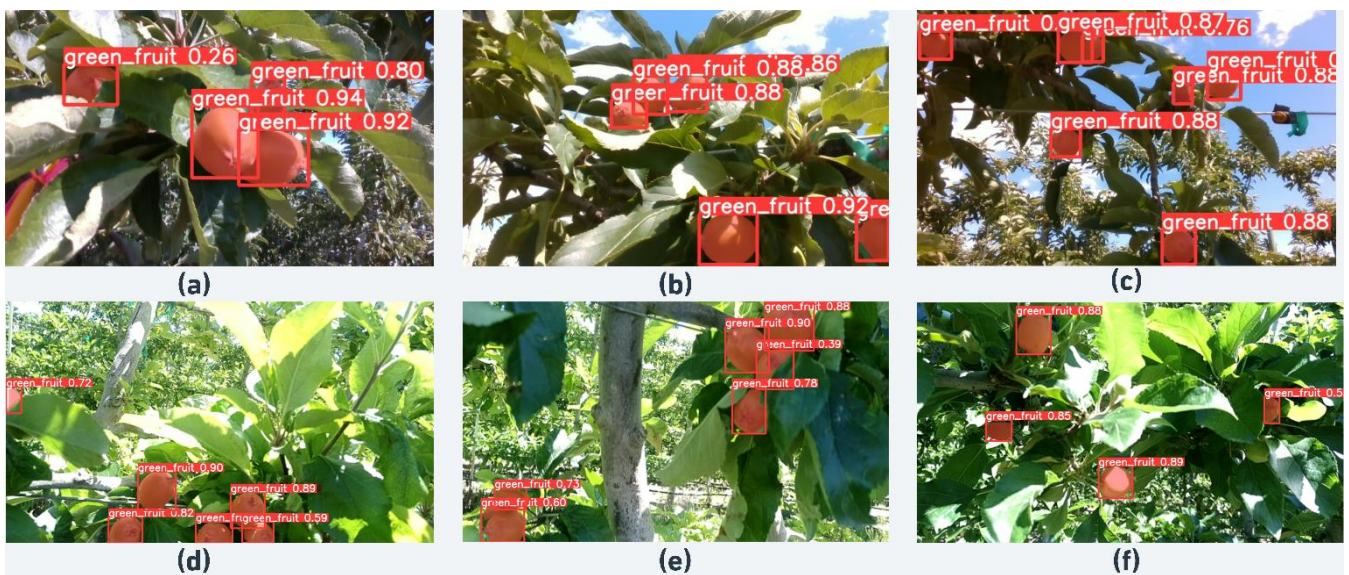

*Figure 6: YOLOv8 detection and segmentation results illustration; (a, b and c) segmentation examples of occluded immature apples in the images captured by IntelRealsense D435i in variable lighting condition; and (d, e and f) segmentation results illustration in complex orchard environment in the images captured using Microsoft Azure Camera.*

These challenges included occlusions caused by leaves and low light or shadow conditions. Figure 6a and 6b presented examples of failed detections (or false negatives) due to leaf occlusions on green apples, while Figure 6c illustrated the failure in detecting and segmenting green apples under low light or shadowed conditions in images. Future

Another similar study focused on green fruit segmentation and orientation estimation for robotic green fruit thinning of apples [44]. There has also been an increasing trend towards channel-pruned and optimized versions of YOLO for fruit detection, as highlighted in studies such as [45] and [46].

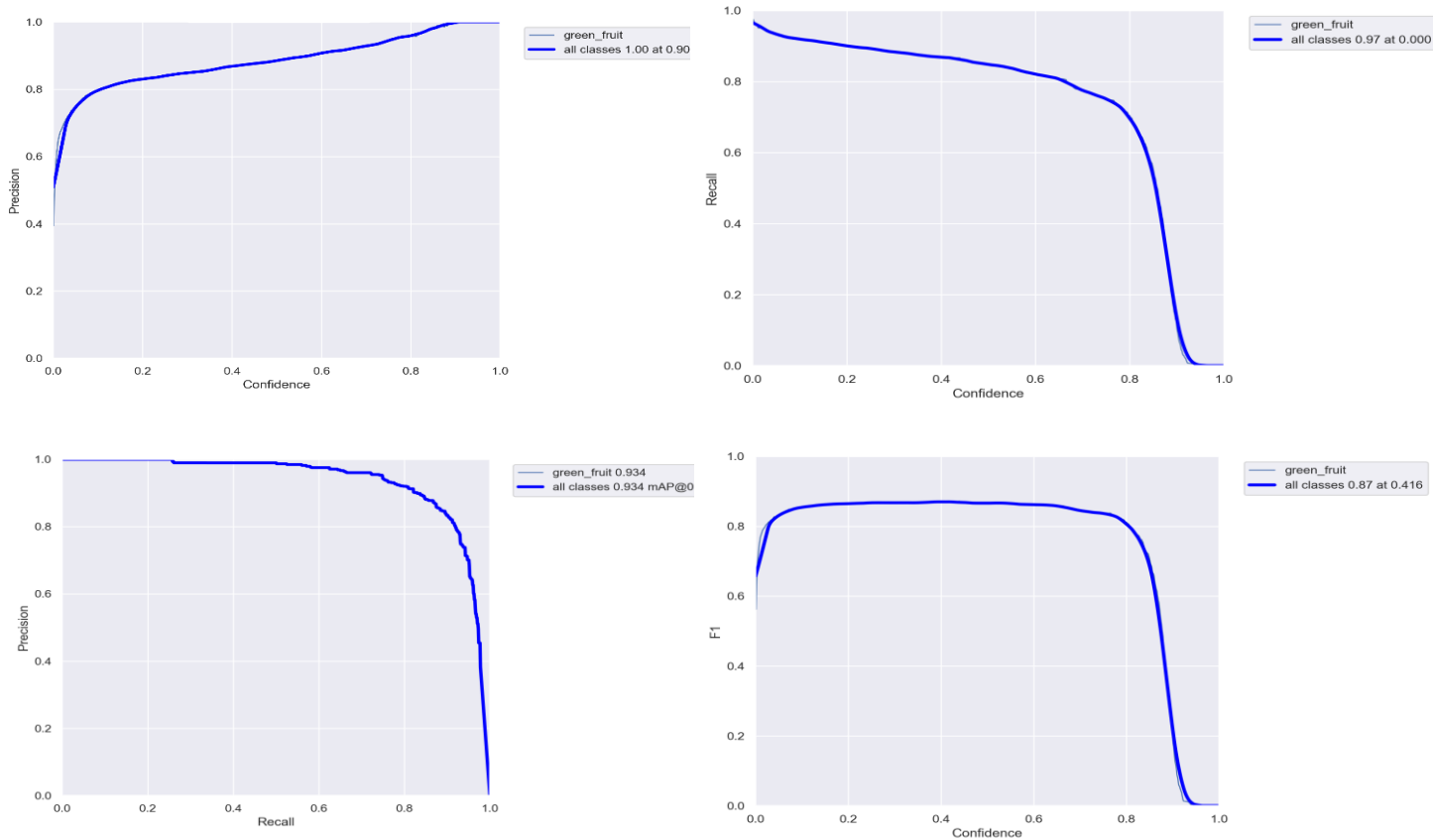

*Figure 7: Immature apples segmentation results achieved with YOLOv8m-seg; (a) Precision-confidence curve; (b) Recall-confidence curve; (c)Precision-Recall Curve; and (d) F1-Confidence curve*

Furthermore, the potential of the YOLOv5-based models has been assessed in detecting litchi fruits for yield estimation [47],spikelet detection in grapes [48] and green pepper detection [49].

While these studies made significant strides in their respective domains, they predominantly utilized standard RGB images or those captured using non-specialized machine vision sensors. Furthermore, most of the existing studies were performed under indoor/greenhouse agriculture. In contrast, the research presented here was based on data collected from a commercial orchard using two of the most widely used sensors in agricultural automation studies. In addition, the immature green fruit detection results obtained in this study with the YOLOv8 model surpassed those reported in the aforementioned studies, particularly in terms of processing speed and accuracy. As illustrated in Figure 8a, an immature apple partially obscured by overlapping leaves and branches was not detected by the model. Similarly, Figure 8b and 8c depicts an interference from neighboring fruits and foliage, which also results in a false negative. Such occlusions can not only introduce inaccuracies in the segmentation mask but also subsequently affect the precise estimation of the green apple sizes.

Training with a higher number of data samples with more occlusion examples could possibly increase the detection and segmentation of immature apples in such conditions.

### B. Immature Apple (Fruitlet) Size Estimation
The ellipsoid fitting technique yielded optimal results in estimating immature apple size in terms of RMSE, MAE, MAPE, and R-squared values. This technique achieved an RMSE of 2.35 mm, MAE of 1.66 mm, MAPE of 6.15 mm, and an R-squared value of 0.9 for images taken with the Microsoft Azure Kinect camera. Similarly, the same technique achieved an RMSE of 9.65 mm, MAE of 7.8 mm, MAPE of 29.48 mm, and an R squared value of 0.77 for IntelRealsense images. Figure 9a and 9b visualized the diameter estimates for 102 immature green apple samples achieved with the three employed shape fitting techniques on images from Microsoft and IntelRealsense cameras, respectively. The performance metrics—RMSE, MAE, MAPE, and R-squared—obtained from the three techniques on both sets of camera images are also presented in Figure 10. The indoor sizing test (Figure 5a) conducted using synthetic green apples with varying sizes, including diameters of 24mm, 27mm, 30mm, and 70mm also showed that the combination of Microsoft Azure with ellipsoid fitting approach achieved maximum accuracy. The RMSE values for the size estimation using Least Square-based sphere fit, RANSAC-based sphere fit, and Ellipsoid fit for the images captured with the Intel RealSense camera were 32.11 mm, 50.22 mm, and 15.36 mm, respectively. In contrast, when utilizing the Microsoft Azure Kinect camera, the corresponding RMSE values were found to be 6.21 mm, 8.74 mm, and 5.74 mm, respectively. Overall, the model achieved better performance on images acquired with Microsoft Azure camera in terms of predicting size using the all three methods tested in this study. A high R-squared value signified that the ellipsoid model could account for 90% of the variance in the given actual apple sizes, emphasizing its precision and reliability in capturing the true dimensions of immature green apples in 3D space. It is noted that the segmentation model was trained using equal proportion of images from both sensors and it was assumed that the sizing difference was primarily caused by the difference in quality and density of 3D point clouds generated by the two sensors used. Future studies could be conducted to further quantify the sources of errors causing the performance differences between these two sensors.

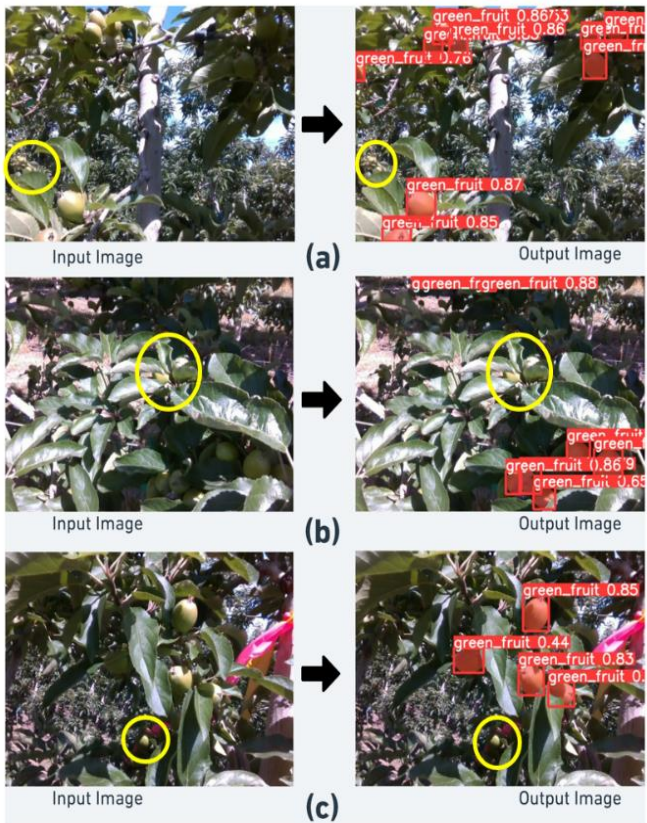

*Figure 8: Occlusion examples. Yellow region indicates the affected region; (a) showing occlusion due to leaves; (b) Occlusion due to leaves and stem; and (c) Foliage and shadow effect resulting in failure in detection.*

In an outdoor orchard environment, our study faced several challenges, particularly due to the occlusion of target objects and diverse environmental factors including variable lighting conditions. The fruit tree canopies included dense foliage and widely overlapping leaves and branches, often obscuring the immature green apples from the sensors' view. This occlusion led to instances of missed detections or inaccurate segmentations. Additionally, variable lighting conditions, such as strong sunlight or shadows, further complicated the detection process. It is also noted that the generalization of the model is hampered by the variability in shape, size, color, texture, and other geometric and spectral characteristics of objects of interest such as immature fruit and leaves.

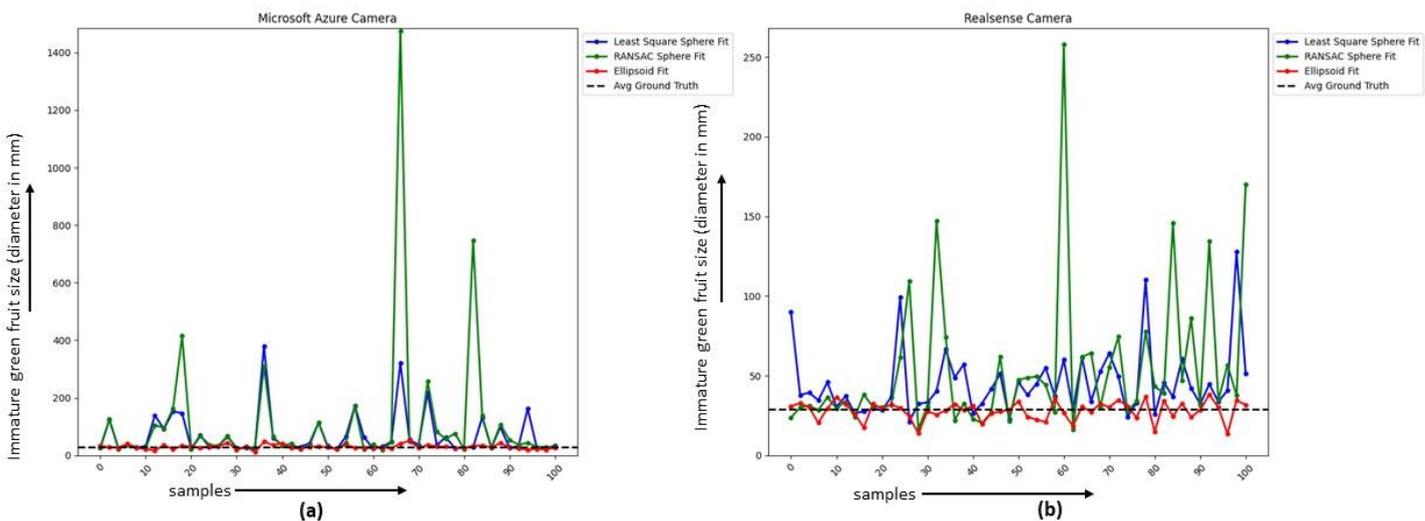

*Figure 9: Results of immature green apple size estimation using Least Square sphere Fitting, RANSAC sphere fitting and Ellipsoid fitting on the images collected using; (a) Microsoft Azure Kinect DK camera; and (b) IntelRealsense d435i camera.*

Our analysis also showed a good correlation between immature apple size and model confidence in estimating it. Larger immature apples, typically exceeding 30 mm in diameter, yielded higher confidence scores, averaging around 0.92, due to their clear visibility and distinct contours. In contrast, the size of smaller or less mature apples under 24 mm were estimated with a much lower average confidence score of ~ 0.75. This reduction in the confidence was caused also by their appearance, which tended to blend with the surrounding foliage when the fruit were in their early stage of development. This variation in confidence scores with apple size and maturity highlights the model's variable sensitivity across different stages of fruit development.

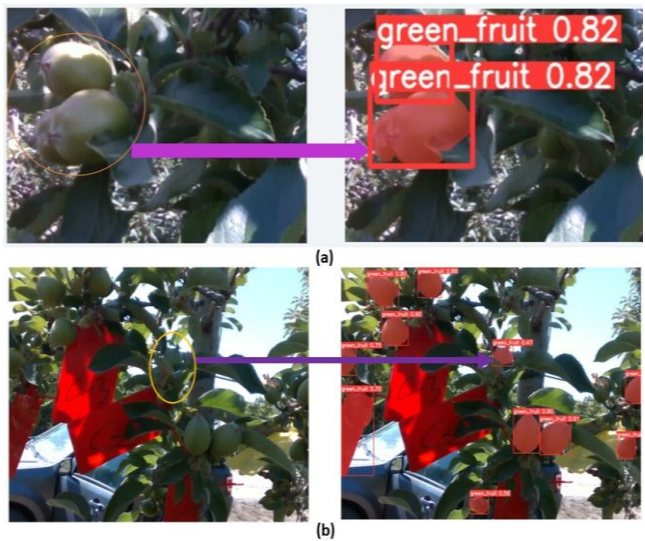

*Figure 10: Over-masking and under-masking during the YOLOv8 segmentation of immature green apples causing errors in estimating their size; (a) showing two immature apples segmented as one immature apple, and (b) only the top part of the immature apple is segmented.*

Figure 10 depicts a scenario where the YOLOv8 algorithm resulted in over-masking and under-masking in the segmentation of immature apples. Figure 10 a illustrates a condition where two immature apples are segmented as a single immature apple. This resulted in a bigger mask than the actual mask in 3D image processing, thus affecting the size estimation of the immature apple. Likewise, Figure 10b demonstrates a condition where the immature apple is occluded by the leaves and foliage, which is a major challenge in agricultural computer vision. Since only half (approximate) part of the immature apple has been segmented by the YOLOv8 deep learning model, the extracted point clouds could not be enough to fit the shape to estimate the size of the apple. Figure 11 shows the RMSE, MAE, MAPE, and R-squared distribution achieved using the three shape-fitting techniques on the images collected using the Microsoft Azure camera and IntelRealsense D435i camera.

Building on the foundational insights offered by prior studies, our investigation ventured further into the realm of 3D imaging and object sizing. For instance, studies like [50], deployed calibration spheres on trees, using them as reference scales to gauge segmented apple sizes. Likewise, 3D solutions have been investigated to reconstruct objects using multi-sensor inputs [51]. Despite their innovative approaches, these methods face challenges with computational demands and occlusions, leading to incomplete reconstructions. Subsequent solutions by [52]and [53] investigated automated shape completion, fitting ellipsoids to gathered point clouds. Yet, they faced challenges due to the plant's dynamic structure or the computational load. Furthermore, [54] explored 3D sizing by aligning an apple's major axis with 3D points from a solitary camera shot combined with a time-of-flight sensor.

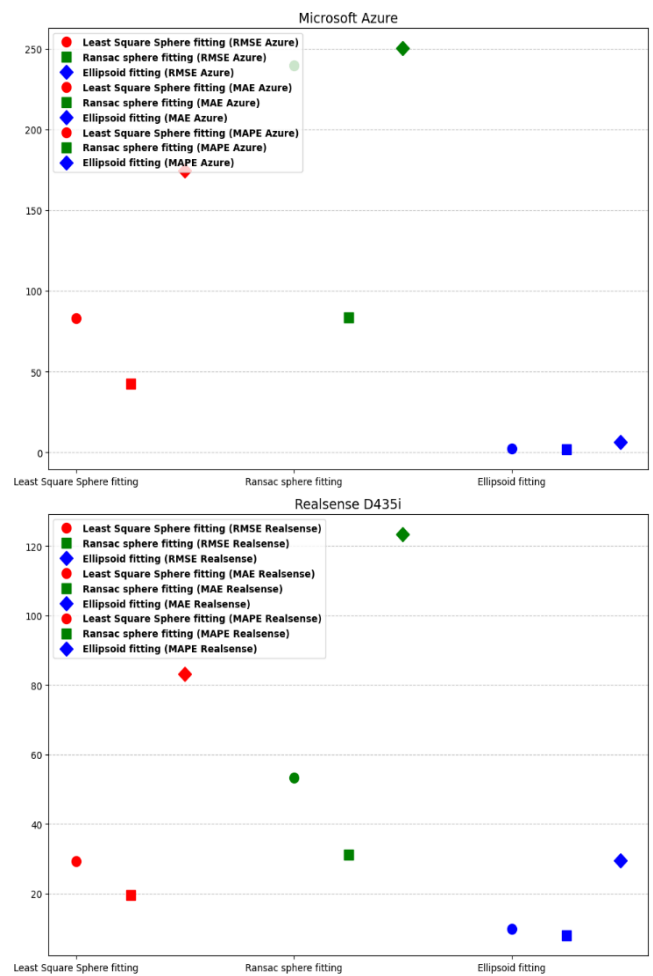

*Figure 11: RMSE, MAE, MAPE, and R-squared distribution achieved using the 3 shape fitting techniques on the images collected using Microsoft Azure camera and IntelRealsense D435i camera.*

They reported an accuracy of 69.1%, which has a big room for improvement. Similarly, [55] sized apples from singular images by fitting 3D spheres, however, this technique is not suited for immature apples given their small and irregular shape. However, these novel strategies encountered hurdles, predominantly due to computational demands and occlusions, culminating in partial reconstructions.

In the process of ellipsoid fitting, an ellipsoidal model is mathematically fitted to the 3D point cloud data of segmented immature apples, capturing their often-elongated shape more accurately than spherical models. This method consistently outperformed sphere fitting as it better accommodates the variability in fruit shapes, especially in early growth stages where apples are not perfectly spherical. Additionally, the YOLOv8 model training, vital for accurate segmentation, involved fine-tuning various parameters such as a learning rate set at 0.001 and the batch size of 32. These settings, combined with the model's anchor-free detection approach, contributed significantly to its high precision and efficiency in complex orchard environments.

It is noted that the YOLOv8 model's anchor-free approach plays a pivotal role in precisely and efficiently detecting objects by directly estimating their centers rather than relying on predefined anchor boxes, thus enhancing the accuracy and speed of immature apple segmentation in complex orchard environments.

For this application, the single-shot detection capability of YOLOv8 is highly advantageous, as it effectively processes predictions in complex scenarios, such as distinguishing between green immature apples (fruitlets) and the similarly colored tree foliage. In comparison, other models like Mask R-CNN, especially in segmentation tasks, tend to have slower computational speeds both during training and real-time field applications (inference), making YOLOv8 a more efficient choice for this context.

This study fills a research gap in detecting and sizing immature apples in commercial orchards, an area not extensively covered in current literature. While we found some related studies as discussed in our literature review such as the use of Mask R-CNN, U-Net, YOLOv5, and YOLOv4, our approach using the YOLOv8 model and 3D shape fitting methods is novel in solving this unique problem of early-stage apple sizing.

## V. DISCUSSION
This study on detection and sizing of early-stage, immature apples was underscored by the critical need in orchard operations, particularly for yield prediction, optimization of crop load management during the growing season, and post-harvest management including storage and marketing. The advent of labor shortages, notably intensified by the COVID-19 pandemic, has highlighted the need for automated solutions in orchard crops. This study employed YOLOv8 models and 3D shape-fitting techniques for precise detection and measurement of apple sizes during their early growth phases, which is important to predict crop-load and fruit distribution in the tree canopies as early in the season as possible. These models demonstrated a potential in automating the detection process in crop-load management (e.g., green fruit thinning) with a practically applicable accuracy, which will significantly reduce the dependency on manual labor in crop-load management in apple orchards. The application of 3D shape-fitting techniques further improved the accuracy of size estimation, employing sphere and ellipsoid models to represent the 3D shape of immature apples accurately. The ellipsoid fitting method, in particular, was found to be more congruent with the natural shape of the fruit in the early growth stages, outperforming traditional spherical models in capturing the true dimensions of the apples. As mentioned before, this approach offered a precise estimation of fruit size and showcased the potential of enhancing automated crop monitoring and management in orchards.

## VI. CONCLUSION AND FUTURE SUGGESTIONS
Detecting and estimating the size of immature apples during their early growth stages is critically important for numerous agricultural processes, from predicting yield and market reception to making informed decisions about crop load and performing robotic green fruit thinning. Traditional methods, while effective, are labor-intensive and the recent labor shortages, exacerbated by the COVID-19 pandemic, underscore the pressing need for automated solutions. To address this challenge, the present study focused on detecting and sizing apples in early growth stages using YOLOv8 models and 3D shape-fitting techniques.

The key findings of this study are:

- The YOLOv8 object detection model demonstrated superior performance in the detection and segmentation of immature green apples, achieving an accuracy of 94%.
- The Ellipsoid fitting method consistently achieved higher accuracy and efficiency in estimating fruit sizes over other techniques tested, with an R-squared value of 0.9 on images acquired with Microsoft camera. The same with Sphere fitting was 0.77.
- The immature apple sizing technique achieved better results for the images acquired with Microsoft Azure Kinect DK sensor compared to the same acquired with Intel RealSense 435i. It was found that RMSE of 2.35 was achieved for images acquired with Azure Kinect whereas the same was 9.65 for images acquired with Realsense.

The automation of green fruit detection and sizing not only addresses current labor challenges but also holds the potential for substantial cost savings and improved crop quality. For future research endeavors, it would be beneficial to expand the dataset and explore other advanced machine-learning algorithms to minimize the impact of variable orchard conditions and occlusion caused by leaves, branches, and other fruit. A multi-sensor fusion approach and imaging with multiple viewing angles might also help improve the robustness of the system and make it applicable to other crops as well.

It is noted that to address the labor-intensive and laborious nature of traditional fruit sizing methods, this study leveraged the YOLOv8 model and 3D shape fitting to offer an automated, accurate, and rapid alternative, thus significantly reducing manual labor while enhancing the efficiency of sizing immature green apples in orchard environments. Automating this process using consumer-grade cameras increased the commercial viability of this technique, which in the future is expected to reduce the dependence on manual labor in sizing immature green fruits in fruit crop production.

In the future, the outcome of this study could be deployed into practical applications in apple orchards. The primary aim is to utilize the YOLOv8 model and 3D shape-fitting techniques for tasks such as automated green fruit thinning and early yield prediction, which would be facilitated by the accurate detection and sizing of immature green apples. For automated green fruit thinning, an embedded computational hardware may have to be integrated with a robotic system for implementing YOLOv8 model. For applications such as early crop prediction, small sensing module could be developed using RGB-D cameras and image acquisition interface to collect images, which could then be processed using in-house or cloud computing platforms. The results could be presented to growers for making informed decisions using user-friendly software interfaces that may run in web or mobile Applications.

Note: The published version of this article can be found on IEEE Access as referenced by [56].

## VII. ACKNOWLEDGEMENT
This research is funded by the National Science Foundation and United States Department of Agriculture, National Institute of Food and Agriculture through the "AI Institute for Agriculture" Program (Award No. AWD003473). The authors gratefully acknowledge Dave Allan (Allan Bros., Inc.) for providing access to the orchard during the data collection and field evaluation.

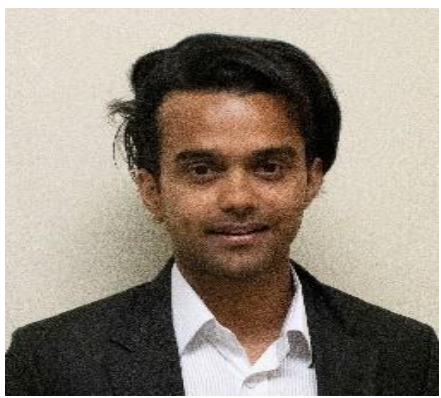

**Ranjan Sapkota** obtained his B.Tech degree in Electrical and Electronics Engineering from Uttarakhand Technical University, India in 2019. He then pursued his MS in Agricultural and Biosystems Engineering from North Dakota State University from 2020 to 2022. Currently, Mr. Sapkota is a Ph.D. student at Washington State University in the Department of Biological Systems Engineering and Center for Precision and Automated Agricultural Systems. His research primarily focuses on Automation and Robotics for Agriculture, utilizing Artificial Intelligence, Machine Vision, Sensing Technologies, Robotic Motion Planning Systems, Robot Manipulation Systems, and Generative AI.

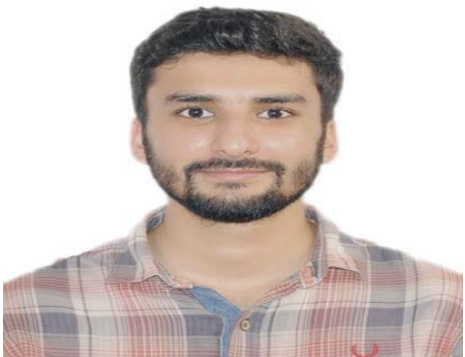

**Dawood Ahmed** obtained his Bachelor's degree in Mechatronics Engineering from the National University of Science and Technology, Pakistan. He is currently pursuing a Ph.D. in Biological Systems Engineering at Washington State University Center for Precision and Automated Agricultural Systems, where he is dedicated to developing robotics and automated solutions using machine vision and deep learning technologies. Mr. Ahmed's research interests include the integration of automation and robotics in various fields, including agriculture, with a particular focus on the application of machine vision and deep learning technologies.

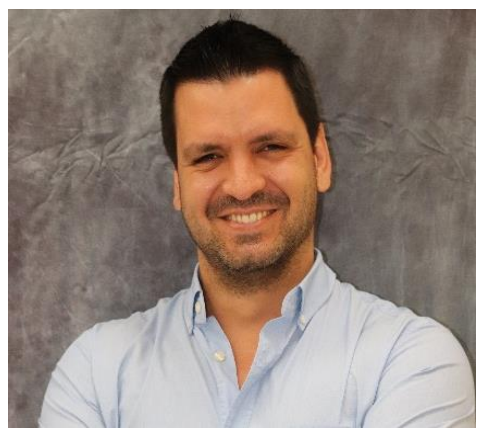

**Martin Churuvija** obtained his Bachelor's degree in Electronics Engineering with a major in Mechatronics/Control from the Buenos Aires Technological Institute, Argentina in 2012. He is a co-founder of Frutobotics, a company that utilizes mechatronic technology to increase the efficiency of harvest and other labor-intensive tasks involved in the production of tree fruits for the fresh market. Churuvija is currently pursuing a PhD degree in Biological and Agricultural Engineering at Washington State University Center for Precision and Automated Agricultural Systems, where he focuses on developing machine vision-based robotic solutions to automate pruning for sweet cherry and apple trees grown in fruiting wall canopy architectures.

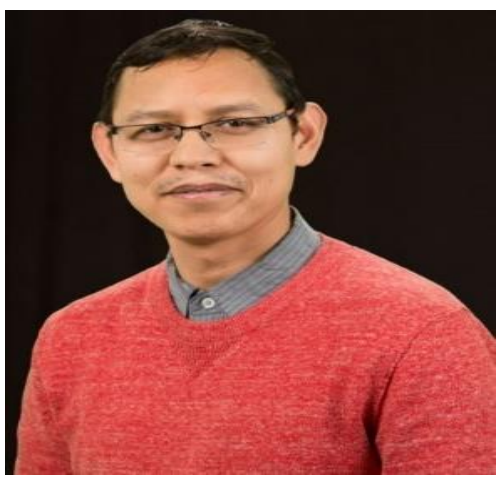

**Prof. Dr. Manoj Karkee** obtained his BS in Computer Engineering from Tribhuwan University in 2002. He pursued his MS in Remote Sensing and Geographic Information Systems at Asian Institute of Technology, Thailand, and earned his Doctorate in Agricultural Engineering and Human-Computer Interaction from Iowa State University in 2009. Dr. Karkee currently serves as the Professor and Director of the Center for Precision and Automated Agricultural Systems at Washington State University. His research focuses on agricultural automation and mechanization programs, with an emphasis on machine vision-based sensing, automation and robotic technologies for specialty crop production.